# An Agent-Based Modeling Approach to Free-Text Keyboard Dynamics for Continuous Authentication

Roberto Dillon* and Arushi

James Cook University, 149 Sims Drive 387380, Singapore

roberto.dillon@jcu.edu.au, arushi@my.jcu.edu.au

https://orcid.org/0000-0003-0166-0273; https://orcid.org/0000-0002-1877-9978

*Corresponding author

**Abstract**

Continuous authentication systems leveraging free-text keyboard dynamics offer a promising additional layer of security in a multifactor authentication setup that can be used in a transparent way with no impact on user experience. This study investigates the efficacy of behavioral biometrics by employing an Agent-Based Model (ABM) to simulate diverse typing profiles across mechanical and membrane keyboards. Specifically, we generated synthetic keystroke data from five unique agents, capturing features related to dwell time, flight time, and error rates within sliding 5-second windows updated every second. Two machine learning approaches, One-Class Support Vector Machine (OC-SVM) and Random Forest (RF), were evaluated for user verification. Results revealed a stark contrast in performance: while One-Class SVM failed to differentiate individual users within each group, Random Forest achieved robust intra-keyboard user recognition (Accuracy > 0.7) but struggled to generalize across keyboards for the same user, highlighting the significant impact of keyboard hardware on typing behavior. These findings suggest that: (1) keyboard-specific user profiles may be necessary for reliable authentication, and (2) ensemble methods like RF outperform One-Class SVM in capturing fine-grained user-specific patterns.

## 1. Introduction

As the modern working environment continues to evolve and faces new threats [1], authenticating users becomes increasingly critical, yet traditional methods may prove insufficient. While passwordless approaches like Passkeys, which merge two-factor authentication by combining biometric verification (i.e. "something we are") on a specific device followed by a device-based authentication (i.e. "something we have") are gaining traction, passwords (i.e. "something we know") remain prevalent due to their simplicity and scalability. However, as username/password pairs grow vulnerable and standard 2FA risks user friction in frequent login scenarios, a transparent secondary authentication layer leveraging behavioral analytics emerges as an alternative solution worth exploring.

Keyboard dynamics refers to the behavioral biometric modality that analyzes unique typing patterns, including keystroke rhythm, latency between key presses, and hold durations [2]. Rooted in telegraph operators' ability to recognize colleagues through Morse code rhythms in the 19th century [3], this approach can provide a novel computational method for user authentication [4] and even for continuous authentication where, unlike static credentials, keyboard dynamics operates in the background by creating a biometric profile for a specific user based on habitual typing characteristics [5]. Existing studies predominantly focus on fixed-text authentication, where users type predefined phrases (e.g., passwords) while free-text analysis, which is critical for authenticating real-world applications like email composition or document editing, remains relatively under-explored [6]. This gap stems partly from the scarcity of public, labeled datasets capturing diverse typing behaviors in an unrestricted context that matches the researchers' specific approaches [7]. Without such data, developing robust models to distinguish legitimate users from impostors to identify possible Business Email Compromise (BEC) attacks, for example, proves challenging.

To address data limitations, this study proposes an agent-based modeling (ABM) framework simulating different users' free-text typing behaviors where agents allow for detailed customization including not only an average typing speed and simulated error rate, like in [8], but also a realistic typing frequency dependent of language (e.g. American English) that takes into account key distances and a dominant hand (right vs left handed typist) as well as a fatigue factor. By training models via simulated agents, it is possible to quickly build diverse datasets to experiment with for detecting anomalies indicative of account takeover attempts, while also circumventing privacy concerns associated with real-user data.

## 2. Agent Based Modeling

A comprehensive agent-based model (ABM) was developed to simulate realistic typing behaviors across diverse user profiles. The typing process involves complex biomechanical interactions between a user's typing habits and the physical constraints of keyboard layouts. Our model conceptualizes typing as a sequence of discrete keystroke events characterized by two key temporal metrics: dwell time (key press duration) and flight time (delay between consecutive keys). We posit that realistic flight times must account for the physical keyboard geometry, individual motor patterns, and physiological factors such as fatigue. Each typing agent in our model represents a unique user with configurable parameters that determine their typing behavior. The agent architecture follows an event-driven paradigm where keystroke timing emerges from the interaction of multiple underlying factors. The system, in fact, maintains internal state variables including current fatigue level, typing speed, and memory of previously typed characters to inform subsequent keystroke timing decisions (e.g. when the agent is simulating the repetition of the same alphabet letter, flight time will be drastically reduced).

The ABM implementation consists of several interconnected components:

- **Text Generation System**: Produces English-like character sequences using a probabilistic model based on character frequency distribution in English text, where character $c$ is sampled according to: $P(c) = \frac{f_c}{\sum_i f_i}$ where $f_c$ represents the frequency of character $c$ in English text. English letter frequency is normalized to sum = 1.0 and was derived from empirical linguistic studies [9].
- **Keyboard Geometry Model**: Maps each key $k$ on a QWERTY layout to spatial coordinates $(x_k, y_k)$, where x and y represent the row and column on a standard keyboard, e.g. ``` `` ``` is the key at position (0,0) in an American keyboard layout, enabling calculation of precise Euclidean distances between any key pair.
- **Keyboard Type**: Two types of keyboards are possible: laptop (membrane) and mechanical. The first is identified by a shorter default dwell time but a longer flight time (50ms vs 60ms and 120ms vs 100ms, respectively) as noted in [10].
- **Personal Pattern Generator**: Creates unique typing fingerprints for each agent through randomized but consistent variations in key-pair transition speeds. Previous simulations like [8] typically used uniform random distributions for flight times regardless of key positions. Here, instead, our approach calculates flight times between any two keys $(k_i, k_j)$ through the following mathematical formulation:

$$F_{i,j} = B \cdot \left(0.5 + \frac{D_{i,j}}{2}\right) \cdot P_{i,j} \cdot (1 + 0.4 \cdot \phi^2) \cdot N \qquad (1)$$

Where:

- $F_{i,j}$ is the flight time between keys $k_i$ and $k_j$ (in milliseconds)
- $B$ represents the base flight time determined by keyboard type: laptop or mechanical.
- $D_{i,j}$ is the normalized Euclidean distance between keys, calculated as: $D_{i,j} = \sqrt{(x_i - x_j)^2 + (y_i - y_j)^2}$
- $P_{i,j}$ captures the user's individual proficiency with the specific key transition, which is affected by whether the simulated user is right or left-handed.
- $\phi$ is the current fatigue level (0-1 range)
- $N$ represents natural timing noise, where $N \sim \mathcal{N}(1, 0.05)$

- **Fatigue Simulator**: Models the non-linear degradation of typing performance over extended sessions using quadratic fatigue accumulation functions.

- **Error and Correction Mechanism**: Simulates realistic typing errors with probability $e$ per character and subsequent backspace correction behaviors, with error correction sequences modeled as: $t_{backspace} \sim \mathcal{N}(40, 3)\ ms$ so that backspaces are simulated with shorter dwell times (μ=40 ms vs. μ=50–60 ms for regular keys) to reflect rapid corrections [10].

Then, to capture the distinctive "signature" of individual typing behaviors, the model implements a personal variation matrix. This matrix assigns efficiency multipliers to each possible key transition pair to simulate specific digraphs. The personal factor for any key pair ($P_{i,j}$) is calculated as:

$$P_{i,j} = V_{base} \cdot V_{hand} \cdot V_{digraph} \qquad (2)$$

Where:

• $V_{base} \sim \mathcal{N}(1.0, 0.15)$ represents random baseline variations, bounded to [0.7, 1.3]

• $V_{hand}$ is the hand dominance factor equal to 0.9 if the keys are close to the dominant hand side (right/left-handed typist), 1.05 if close to the non-dominant hand, and 1.0 otherwise.

• $V_{digraph} = 0.85$ if the key pair forms a common English digraph, 1.0 otherwise.

For repeated characters (when $k_i = k_j$), we use a special case of $D_{i,j} = 0.2$ to reflect the significantly reduced movement time. The resulting pattern creates a unique typing fingerprint for each agent while maintaining realistic constraints based on keyboard ergonomics and linguistic patterns. Last, a distinctive feature of our model is also the incorporation of realistic fatigue effects that accrue during extended typing sessions. The fatigue accumulation follows:

$$\phi_{t+1} = \min(1.0, \phi_t + \gamma) \qquad (3)$$

Where $\phi_t$ is the fatigue level at time $t$ and $\gamma$ is the user-specific fatigue factor. This accumulated fatigue affects the current typing speed through a quadratic relationship to create a more realistic fatigue progression, with minimal impact initially but accelerating deterioration as typing continues:

$$WPM_t = WPM_{base} \cdot (1 - 0.3 \cdot \phi_t^2) \qquad (4)$$

The system was implemented in Python, using NumPy for mathematical operations and random sampling. The agent's state is fully encapsulated, allowing for the concurrent simulation of multiple users with distinct characteristics. The output consists of time-stamped keystroke events (both press and release actions) that record the exact timing of each interaction with millisecond precision.

## 3. Dataset Generation

A specific user generated by the ABM is identified by the following parameters:

- agent_id (int),
- wpm (float,e.g. 45.0),
- error_rate (float, e.g., 0.5),
- keyboard_type (str, e.g. "laptop" or "mechanical"),
- fatigue_factor (float, e.g. 0.001),
- finger_agility (float, e.g. 1.0)
- dominant_hand (str, e.g. "right" or "left").

Five different users were defined, with different typing characteristics (Table 1):

Table 1. User profiles simulated by ABM. Specific values per user were selected within predefined ranges at runtime. Note that User 2 is more prone to fatigue compared to the others while User 1 is the only left-handed among the group.

| id | WPM. Range: | Error rate. Range: | Fatigue factor. Range: | Finger agility. Range: | Dominant hand |
|----|-------------|--------------------|-----------------------|------------------------|---------------|
| 1  | 50.0 - 55.0 | 0.03 - 0.04        | $1 \times 10^{-4}$ - $3 \times 10^{-4}$ | 0.9 - 1.0 | Left  |
| 2  | 65.0 - 70.0 | 0.01 - 0.03        | $1.5 \times 10^{-3}$ - $3 \times 10^{-3}$ | 1.0 - 1.1 | Right |
| 3  | 40.0 - 45.0 | 0.02 - 0.03        | $1 \times 10^{-4}$ - $3 \times 10^{-4}$ | 0.8 - 0.9 | Right |
| 4  | 80.0 - 85.0 | 0.08 - 0.10        | $1 \times 10^{-4}$ - $3 \times 10^{-4}$ | 1.2 - 1.3 | Right |
| 5  | 30.0 - 35.0 | 0.01 - 0.02        | $1 \times 10^{-4}$ - $3 \times 10^{-4}$ | 0.7 - 0.8 | Right |

For each user, two files were generated simulating the typing of 1000 characters each across the two possible keyboards (i.e. laptop and mechanical) for a total of four files per user. A sample of the data generated is shown in Table 2.

Table 2. Sample data generated by the ABM (User 2 typing on a laptop keyboard)

| Timestamp (ms) | Key | Action | Keyboard type | Agent id | wpm | Error rate | Fatigue factor | Finger agility | Dominant hand |
|----------------|-----|--------|---------------|----------|-----|------------|----------------|----------------|---------------|
| 0              | o   | press  | laptop        | user2    | 65.31 | 0.0293   | 0.002217       | 1.05           | right         |
| 53.37          | o   | release| laptop        | user2    | 65.31 | 0.0293   | 0.002217       | 1.05           | right         |
| 174            | t   | press  | laptop        | user2    | 65.31 | 0.0293   | 0.002217       | 1.05           | right         |
| 220.82         | t   | release| laptop        | user2    | 65.31 | 0.0293   | 0.002217       | 1.05           | right         |

## 4. Feature Extraction

The data generated by the ABM is elaborated to extract the average and standard deviation of both the Dwell Time, defined as the time in ms between a key-press event and its following key-release event, and the Flight Time, defined as the time between a key-release event and the following key-press event, across a sliding time window spanning the last five seconds and updated every second. The Error Rate, defined as the ratio between the number of backspaces and the total number of characters times 100 across the time window, is also included for a total of five features. Table 3 shows a sample of the extracted features for User 4.

**Table 3.** Features extracted from the first 10 seconds of simulated typing for User 4 on a mechanical keyboard.

| Window start_ms | Window end_ms | avg_dwell time | std_dwell time | avg_flight time | std_flight time | Error rate |
|---|---|---|---|---|---|---|
| 0 | 5000 | 51.57906 | 10.60823 | 102.1077 | 90.92518 | 37.5 |
| 1000 | 6000 | 51.69848 | 10.45281 | 101.6424 | 82.8035 | 36.36364 |
| 2000 | 7000 | 54.18645 | 9.115519 | 105.0226 | 78.87076 | 22.58065 |
| 3000 | 8000 | 53.77207 | 8.36145 | 119.0034 | 83.61521 | 20.68966 |
| 4000 | 9000 | 53.58179 | 7.938226 | 116.7275 | 74.17365 | 17.85714 |
| 5000 | 10000 | 56.3344 | 4.010859 | 147.3948 | 69.19318 | 0 |

Kolmogorov-Smirnov tests (KS Tests) performed on each feature for each pair of simulated typing sessions show statistically significant differences between users (p < 0.05) and significant similarities among sessions by the same agent on the same keyboard (p > 0.05) for most features. On the other hand, comparing features extracted from testing the different sessions by the same user across the two different keyboards showed extreme differences, with all p values very close to 0.00. See Table 4 for Laptop and Table 5 for Mechanical keyboard, respectively. When comparing the same agent across different typing sessions, one full session is used for training, and it is then tested with the other one. When comparing with different agents, instead, the tables show the average p-values obtained from testing the session from the first agent with both available simulations from the other agent.

**Table 4.** KS Tests performed for each agent pair with the Laptop keyboard setting. p-values for each feature, p < 0.05 suggests different distribution. Note: ad = avg_dwell, sd = std_dwell, av = average_flight, sf = standard flight, er = error rate.

| User | 1 | 2 | 3 | 4 | 5 |
|---|---|---|---|---|---|
| 1 | **ad: 0.2739**<br>**sd: 0.3703**<br>**af: 0.6690**<br>**sf: 0.7539**<br>**er: 0.9677** | ad: 0.0000<br>sd: 0.0001<br>af: 0.0000<br>sf: 0.0000<br>er: 0.3435 | ad: 0.0043<br>sd: 0.0005<br>af: 0.0000<br>sf: 0.0000<br>er: 0.0011 | ad: 0.0000<br>sd: 0.0001<br>af: 0.0000<br>sf: 0.0000<br>er: 0.0000 | ad: 0.2908<br>sd: 0.0000<br>af: 0.0000<br>sf: 0.0000<br>er: 0.0000 |
| 2 | ad: 0.0000<br>sd: 0.0034<br>af: 0.0000<br>sf: 0.0000<br>er: 0.0003 | **ad: 0.00521**<br>**sd: 0.0032**<br>**af: 0.0052**<br>**sf: 0.3521**<br>**er: 0.0013** | ad: 0.0000<br>sd: 0.0019<br>af: 0.0635<br>sf:0.1038<br>er: 0.8346 | ad: 0.0000<br>sd: 0.0000<br>af: 0.0000<br>sf: 0.0000<br>er: 0.0000 | ad: 0.0000<br>sd: 0.0002<br>af: 0.0002<br>sf: 0.0005<br>er: 0.6263 |
| 3 | ad: 0.0250<br>sd: 0.0052<br>af: 0.0000<br>sd: 0.0000<br>er: 0.0001 | ad: 0.0000<br>sd: 0.0010<br>af: 0.0344<br>sf: 0.0001<br>er: 0.3762 | **ad: 0.1153**<br>**sd: 0.2399**<br>**af: 0.6099**<br>**sf: 0.0000**<br>**er: 0.5315** | ad: 0.0000<br>sd: 0.0000<br>af: 0.0000<br>sf: 0.0000<br>er: 0.0000 | ad: 0.0182<br>sd: 0.0002<br>af: 0.0007<br>sf: 0.3421<br>er: 0.5139 |
| 4 | ad: 0.0000<br>sd: 0.0000<br>af: 0.0000<br>sf: 0.0000<br>er: 0.0000 | ad: 0.0000<br>sd: 0.0000<br>af: 0.0000<br>sf: 0.0000<br>er: 0.0000 | ad: 0.0000<br>sd: 0.0000<br>af: 0.0000<br>sf: 0.0000<br>er: 0.0000 | **ad: 0.1611**<br>**sd: 0.4153**<br>**af: 0.0000**<br>**sf: 0.0000**<br>**er: 0.2379** | ad: 0.0000<br>sd: 0.0000<br>af: 0.0000<br>sf: 0.0000<br>er: 0.0000 |
| 5 | ad: 0.0000<br>sd: 0.0000<br>af: 0.0000<br>sf: 0.0000<br>er: 0.0000 | ad: 0.0000<br>sd: 0.0002<br>af: 0.0000<br>sf: 0.0004<br>er: 0.1418 | ad: 0.0045<br>sd: 0.0008<br>af: 0.0000<br>sf: 0.2175<br>er: 0.2830 | ad: 0.0000<br>sd: 0.0000<br>af: 0.0000<br>sf: 0.0000<br>er: 0.0000 | **ad: 0.0000**<br>**sd: 0.5415**<br>**af: 0.0441**<br>**sf: 0.1475**<br>**er: 0.7996** |

**Table 5.** KS Tests performed for each agent pair with the Mechanical keyboard setting. p-values for each feature, p < 0.05 suggests different distribution. Note: ad = avg_dwell, sd = std_dwell, av = average_flight, sf = standard flight, er = error rate.

| User | 1 | 2 | 3 | 4 | 5 |
|---|---|---|---|---|---|
| 1 | **ad: 0.0227**<br>**sd: 0.1113**<br>**af: 0.3086**<br>**sf: 0.0154**<br>**er: 0.0669** | ad: 0.0000<br>sd: 0.0000<br>af: 0.0000<br>sf: 0.0000<br>er: 0.0013 | ad: 0.5076<br>sd: 0.3355<br>af: 0.0000<br>sf: 0.0000<br>er: 0.2483 | ad: 0.0 000<br>sd: 0.0000<br>af: 0.0000<br>sf: 0.0000<br>er: 0.0000 | ad: 0.0000<br>sd: 0.0000<br>af: 0.0000<br>sf: 0.0000<br>er: 0.0000 |
| 2 | ad: 0.0000<br>sd: 0.0000<br>af: 0.0000<br>sf: 0.0000<br>er: 0.0297 | **ad: 0.0955**<br>**sd: 0.3074**<br>**af: 0.0371**<br>**sf: 0.0015**<br>**er: 0.0226** | ad: 0.0000<br>sd: 0.0000<br>af: 0.0398<br>sf: 0.1866<br>er: 0.3713 | ad: 0.0000<br>sd: 0.0000<br>af: 0.0000<br>sf: 0.0000<br>er: 0.0000 | ad: 0.0000<br>sd: 0.0000<br>af: 0.0000<br>sf: 0.0511<br>er: 0.0122 |
| 3 | av: 0.2678<br>sd: 0.0000<br>af: 0.0000<br>sf: 0.0000<br>er: 0.0477 | ad: 0.0000<br>sd: 0.0000<br>af: 0.0001<br>sf: 0.0602<br>er: 0.3679 | **ad: 0.1354**<br>**sd: 0.0006**<br>**af: 0.0000**<br>**sf: 0.0000**<br>**er:0.0008** | ad: 0.0000<br>sd: 0.0000<br>af: 0.0000<br>sf: 0.0000<br>er: 0.0000 | ad: 0.0000<br>sd: 0.0000<br>af: 0.0000<br>sf: 0.0177<br>er: 0.0000 |
| 4 | ad: 0.0000<br>sd: 0.0000<br>af: 0.0000<br>sf: 0.0002<br>er: 0.0000 | ad: 0.0000<br>sd: 0.0000<br>af: 0.0000<br>sf: 0.0000<br>er: 0.0000 | ad: 0.0000<br>sd: 0.0000<br>af: 0.0000<br>sf: 0.0000<br>er: 0.0000 | **ad: 0.7486**<br>**sd: 0.5475**<br>**af: 0.4731**<br>**sf: 0.4871**<br>**er: 0.4887** | ad: 0.0001<br>sd: 0.0000<br>af: 0.0000<br>sf: 0.0000<br>er: 0.0000 |
| 5 | ad: 0.0007<br>sd: 0.0000<br>af: 0.0000<br>sf: 0.0000<br>er: 0.0000 | ad: 0.0000<br>sd: 0.0000<br>af: 0.0000<br>sf: 0.0253<br>er: 0.0112 | ad: 0.0017<br>sd: 0.0342<br>af: 0.0180<br>sf: 0.0253<br>er: 0.0035 | ad: 0.0001<br>sd: 0.0000<br>af: 0.0000<br>sf: 0.0000<br>er: 0.0000 | **ad: 0.0001**<br>**sd: 0.3934**<br>**af: 0.0027**<br>**sf: 0.0059**<br>**er: 0.6552** |

## 5. Analysis of Results

Several strategies have been explored in the past showing promising results, from neural networks [6] to OneClass-SVM [8]. In [11], different approaches were tested and compared, with Random Forest showing the best results. Following that example we decided to analyze the ABM dataset via OneClass-SVM and Random Forest to compare the two models that are often reported in literature as being particularly reliable for this and similar contexts.

For both models, one full session per keyboard is used for training, while the remaining sessions served as test sets so as to evaluate the consistency of user-specific typing patterns across sessions and the impact of keyboard variation on model performance.

### 5.1 OneClass-SVM

To evaluate typing consistency and user distinctiveness, a One-Class Support Vector Machine (OC-SVM) was employed as an anomaly detection model. The OC-SVM was trained using the single user's typing data on a specific keyboard (e.g., users 1-5 on simulated laptop keyboard) and evaluated on both the same-user and across users, including variations across keyboard types. The training and testing with OC-SVM allowed us to assess inter- user consistency and inter-user separability, as well as the effect of hardware variation.

Data from each feature was normalized and only the first two principal components obtained via PCA were used as inputs to the One-Class SVM model in order to reduce dimensionality while preserving the majority of the variance in the data. Limiting to two PCA features, besides avoiding overfitting, also enabled a visual distribution of inliers vs outliers which is not possible with higher dimensional data.

The model's performance varied across users and keyboard combinations. In many cases, testing on the same user and same keyboard yielded higher inlier detection rates - for e.g., 78.15% for User 4 and 73.23% for User 5 on the laptop keyboard, and 74.04% for User 2 and 81.25% for User 5 on the mechanical keyboard. Some cross-keyboard evaluations resulted in a higher inlier percentage as well. For example, when trained on laptop and tested on mechanical, the model achieved 72.54% for User 1, 72.31% for User 2, and 79.93% for User 5; similarly, when trained on mechanical and tested on laptop, it achieved 77.85% for User 2, 74.24% for User 4, and 74.15% for User 5. To further assess user distinctiveness, cross-user evaluations were conducted by training one user and testing on others across both keyboards. A summary of inlier detection results across these scenarios is provided in Tables 6 to Table 9. The cross-user inlier detection percentages were consistently high, ranging from 50% to 81.25%, with most values exceeding 60% as shown in Tables 7 and 8. This indicates that the One-Class SVM (OC-SVM) was generally unable to distinguish between different users during testing, often recognizing unfamiliar typing patterns as similar to the training profile.

**Table 6.** Cross-User Inlier Detection Rates Using OC-SVM for Laptop Session 1 for training and Laptop Session 2 for testing.

| Train/Test | User 1 L-2 | User 2 L-2 | User 3 L-2 | User 4 L-2 | User 5 L-2 |
|---|---|---|---|---|---|
| **User 1 L-1** | **65.66** | 53.09 | 71.74 | 70.38 | 69.84 |
| **User 2 L-1** | 60.33 | **69.70** | 61.26 | 67.45 | 59.69 |
| **User 3 L-1** | 59.62 | 71.00 | **55.23** | 62.13 | 60.61 |
| **User 4 L-1** | 78.86 | 74.59 | 82.22 | **78.15** | 83.69 |
| **User 5 L-1** | 69.43 | 78.50 | 78.41 | 71.35 | **73.23** |

Figure 1 illustrates the OC-SVM decision boundary trained on User 5's laptop typing data (User 5 L 1) and tested on User 4's typing on the same keyboard type (User 4 L 2). A large set of test samples fall within the decision boundary, indicating that the model frequently misclassified User 4's typing similar to User 5's typing. The outcome reflects poor user separability and opens further experimentation with OC-SVM for cross-user keystrokes anomaly detection studies.

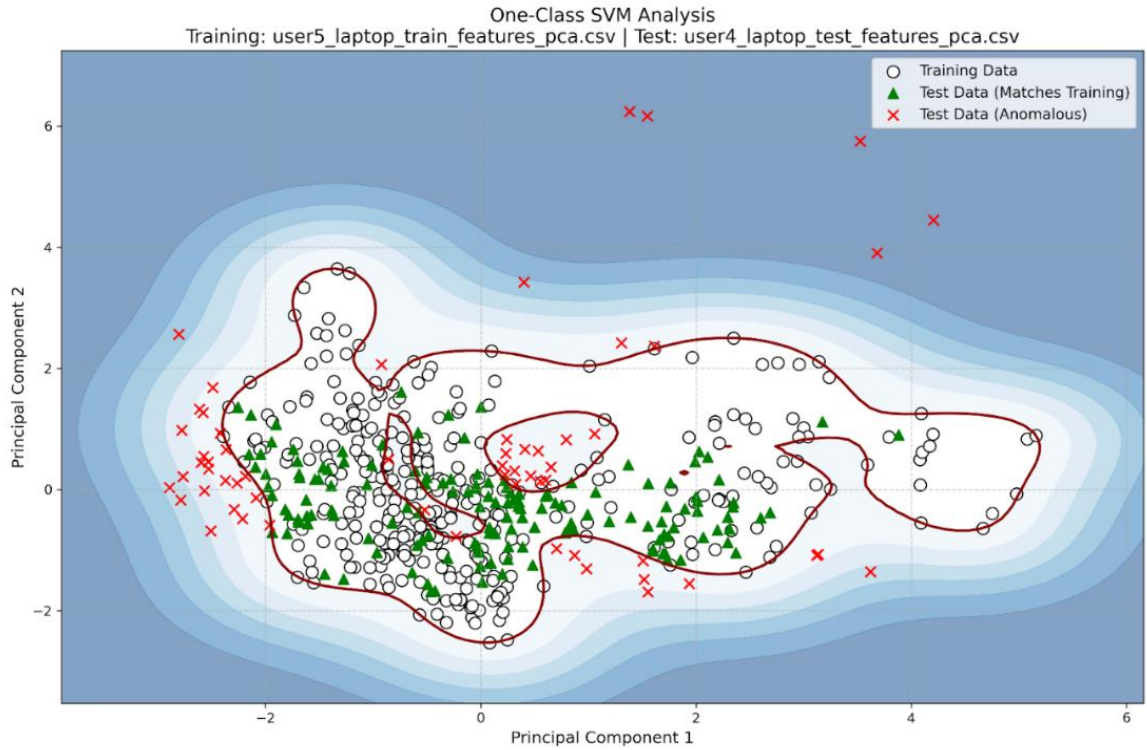

**Figure 1.** Data points for "User 5 L 1" vs "User 4 L 2". More than 70% of the testing data falls in the same region as the original distribution.

**Table 7.** Cross-User Inlier Detection Rates Using OC-SVM for Laptop Session 2 for training and Mechanical Session 2 for testing.

| **Train/Test** | **User 1 M-2** | **User 2 M-2** | **User 3 M-2** | **User 4 M-2** | **User 5 M-2** |
|---|---|---|---|---|---|
| **User 1 L-2** | **72.54** | 58.82 | 68.44 | 56.43 | 76.31 |
| **User 2 L-2** | 66.80 | **72.31** | 61.69 | 63.36 | 64.47 |
| **User 3 L-2** | 58.60 | 73.01 | **55.93** | 50.49 | 61.51 |
| **User 4 L-2** | 74.59 | 73.35 | 75.35 | **68.31** | 77.96 |
| **User 5 L-2** | 77.04 | 82.00 | 77.96 | 62.87 | **79.93** |

**Table 8.** Cross-User Inlier Detection Rates Using OC-SVM for Mechanical Session 2 for training and Laptop Session 1 for testing.

| Train/Test | User 1 L-1 | User 2 L-1 | User 3 L-1 | User 4 L-1 | User 5 L-1 |
|---|---|---|---|---|---|
| **User 1 M-2** | **67.54** | 64.83 | 67.30 | 61.16 | 73.23 |
| **User 2 M-2** | 55.47 | **77.85** | 67.93 | 62.13 | 69.23 |
| **User 3 M-2** | 63.39 | 62.54 | **57.46** | 59.70 | 64.61 |
| **User 4 M-2** | 61.13 | 57.98 | 62.85 | **74.24** | 74.15 |
| **User 5 M-2** | 61.50 | 77.85 | 65.39 | 52.91 | **74.15** |

**Table 9.** Cross-User Inlier Detection Rates Using OC-SVM for Mechanical Session 1 for training and Mechanical Session 2 for testing.

| Train/Test | User 1 M-2 | User 2 M-2 | User 3 M-2 | User 4 M-2 | User 5 M-2 |
|---|---|---|---|---|---|
| **User 1 M-1** | **62.29** | 62.92 | 63.05 | 57.42 | 81.25 |
| **User 2 M-1** | 71.31 | **74.04** | 67.45 | 53.46 | 71.05 |
| **User 3 M-1** | 59.83 | 68.85 | **60.33** | 60.39 | 55.26 |
| **User 4 M-1** | 66.39 | 60.55 | 62.37 | **68.31** | 75.98 |
| **User 5 M-1** | 69.67 | 74.39 | 71.86 | 51.98 | **81.25** |

Figure 2 shows the decision boundary of OC-SVM trained on User 1's mechanical training keyboard data (User 1 M 1) and tested on User 2's testing features on the same keyboard type (User 2 M 2). While some test samples are correctly rejected as outliers, a significant portion of User 2's data still falls within the training boundary, resulting in false acceptance. This highlights the model's limited ability to distinguish between users in cross-user evaluations.

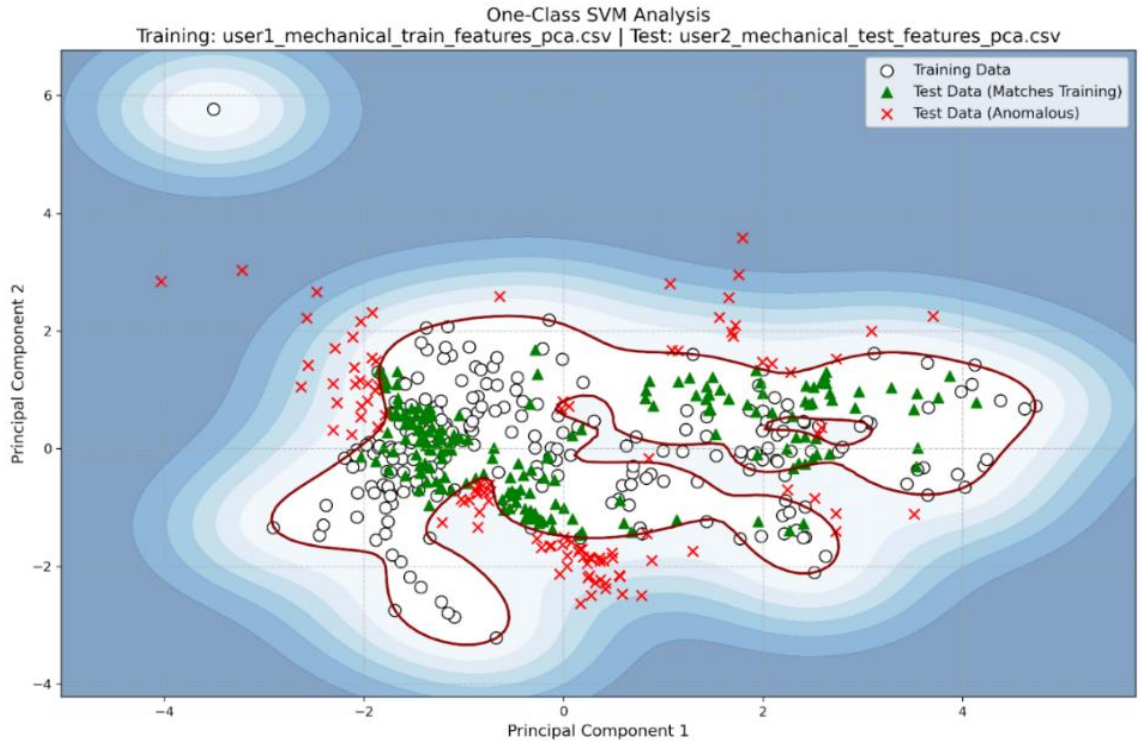

**Figure 2.** Data points for "User 1 M 1" vs "User 2 M 2". More than 60% of the testing data is still recognized as belonging to the original distribution.

### 5.2 Random Forest

The Random Forest (RF) classification algorithm was programmed in Python using the Random Forest Classifier and related functions from the 'scikit-learn'[1] libraries and implemented with the parameters listed in Table 10.

**Table 10.** Random Forest parameters used in the python implementation for analysis

| Parameter | Value | Comment |
|---|---|---|
| n_estimators | 500 | Number of trees in the forest |
| max_depth | 10 | Maximum depth of trees |
| min_samples_split | 5 | Minimum samples required to split a node |
| min_samples_leaf | 2 | Minimum samples required at a leaf node |
| max_features | sqrt | Number of features to consider for best split |
| bootstrap | true | Whether to use bootstrap samples |

Results in terms of overall Accuracy and F1 scores for simulated typing sessions on laptop and mechanical keyboards are summarized in Table 11 and 12 respectively. The Accuracy threshold to decide whether the two distributions, i.e. simulated users, are different, was set to 0.7 (A > 0.70). We classify sessions with a score A < 0.70 as coming from the same user since the two distributions are too similar to be separated consistently. Same agent pairs are highlighted in bold, while misclassified pairs (e.g. different agents recognized as the same agent) are highlighted in bold and *italic*.

---

[1] https://scikit-learn.org/

**Table 11.** Comparison of simulated sessions across laptop keyboards. A = Accuracy. $F_0$ = F1-score for Class 0, $F_1$ = F1-score for Class 1. A > 0.70 indicates the second file can be classified as typed from a different user.

| Laptop | U1-1 | U1-2 | U2-1 | U2-2 | U3-1 | U3-2 | U4-1 | U4-2 | U5-1 | U5-2 |
|---|---|---|---|---|---|---|---|---|---|---|
| User1-1 |  | **A:0.66** $F_0$**:0.65** $F_1$**:0.67** | A:0.88 $F_0$:0.87 $F_1$:0.88 | A:0.90 $F_0$:0.89 $F_1$:0.90 | A:0.71 $F_0$:0.68 $F_1$:0.74 | A:0.78 $F_0$:0.75 $F_1$:0.80 | A:0.84 $F_0$:0.86 $F_1$:0.81 | A:0.76 $F_0$:0.78 $F_1$:0.73 | A: 0.77 $F_0$:0.75 $F_1$:0.80 | A:0.79 $F_0$:0.75 $F_1$:0.82 |
| User2-1 | A:0.86 $F_0$:0.87 $F_1$:0.85 | A:0.91 $F_0$:0.91 $F_1$:0.90 |  | **A:0.67** $F_0$**:0.64** $F_1$**:0.69** | A:0.86 $F_0$:0.87 $F_1$:0.86 | A:0.92 $F_0$:0.92 $F_1$:0.92 | A:0.91 $F_0$:0.93 $F_1$:0.90 | A:0.92 $F_0$:0.94 $F_1$:0.90 | A: 0.93 $F_0$:0.93 $F_1$:0.94 | A:0.93 $F_0$:0.93 $F_1$:0.94 |
| User3-1 | A:0.74 $F_0$:0.77 $F_1$:0.72 | A:0.79 $F_0$:0.81 $F_1$:0.76 | A:0.89 $F_0$:0.88 $F_1$:0.89 | A:0.91 $F_0$:0.91 $F_1$:0.91 |  | **A:0.64** $F_0$**:0.64** $F_1$**:0.63** | A:0.93 $F_0$:0.94 $F_1$:0.9 | A:0.87 $F_0$:0.90 $F_1$:0.83 | ***A: 0.63*** ***$F_0$:0.58*** ***$F_1$:0.67*** | ***A: 0.56*** ***$F_0$:0.59*** ***$F_1$:0.53*** |
| User4-1 | A:0.79 $F_0$:0.76 $F_1$:0.81 | A:0.80 $F_0$:0.77 $F_1$:0.82 | A:0.91 $F_0$:0.92 $F_1$:0.92 | A:0.93 $F_0$:0.91 $F_1$:0.94 | A:0.84 $F_0$:0.82 $F_1$:0.86 | A:0.94 $F_0$:0.92 $F_1$:0.95 |  | **A:0.67** $F_0$**:0.68** $F_1$**:0.65** | A:0.91 $F_0$:0.89 $F_1$:0.92 | A:0.93 $F_0$:0.92 $F_1$:0.95 |
| User5-1 | A:0.86 $F_0$:0.87 $F_1$:0.84 | A:0.81 $F_0$:0.84 $F_1$:0.78 | A:0.94 $F_0$:0.95 $F_1$:0.94 | A:0.93 $F_0$:0.93 $F_1$:0.93 | ***A:0.63*** ***$F_0$:0.64*** ***$F_1$:0.63*** | ***A:0.69*** ***$F_0$:0.71*** ***$F_1$:0.69*** | A:0.92 $F_0$:0.93 $F_1$:0.89 | A:0.98 $F_0$:0.98 $F_1$:0.97 |  | **A:0.56** $F_0$**:0.60** $F_1$**:0.52** |

**Table 12.** Comparison of simulated sessions across mechanical keyboards. A = Accuracy. $F_0$ = F1-score for Class 0, $F_1$ = F1-score for Class 1. A > 0.70 indicates the second file can be classified as typed from a different user.

| Mechanical | U1-1 | U1-2 | U2-1 | U2-2 | U3-1 | U3-2 | U4-1 | U4-2 | U5-1 | U5-2 |
|---|---|---|---|---|---|---|---|---|---|---|
| User1-1 |  | **A:0.68** $F_0$**:0.68** $F_1$**:0.67** | A:0.91 $F_0$:0.91 $F_1$:0.92 | A:0.92 $F_0$:0.92 $F_1$:0.93 | A:0.72 $F_0$:0.72 $F_1$:0.71 | A:0.80 $F_0$:0.78 $F_1$:0.82 | A:0.78 $F_0$:0.81 $F_1$:0.73 | A:0.83 $F_0$:0.85 $F_1$:0.81 | A:0.79 $F_0$:0.75 $F_1$:0.82 | A:0.77 $F_0$:0.72 $F_1$:0.80 |
| User2-1 | A:0.92 $F_0$:0.92 $F_1$:0.91 | A:0.87 $F_0$:0.88 $F_1$:0.86 |  | **A:0.67** $F_0$**:0.67** $F_1$**:0.66** | A:0.91 $F_0$:0.90 $F_1$:0.91 | A:0.92 $F_0$:0.92 $F_1$:0.92 | A:0.95 $F_0$:0.95 $F_1$:0.94 | A:0.90 $F_0$:0.91 $F_1$:0.88 | A:0.91 $F_0$:0.90 $F_1$:0.92 | A:0.90 $F_0$:0.90 $F_1$:0.91 |
| User3-1 | A:0.84 $F_0$:0.86 $F_1$:0.81 | A:0.81 $F_0$:0.83 $F_1$:0.78 | A:0.89 $F_0$:0.90 $F_1$:0.88 | A:0.90 $F_0$:0.91 $F_1$:0.89 |  | **A:0.66** $F_0$**:0.67** $F_1$**:0.66** | A:0.87 $F_0$:0.89 $F_1$:0.84 | A:0.90 $F_0$:0.92 $F_1$:0.88 | A:0.71 $F_0$:0.70 $F_1$:0.74 | A:0.72 $F_0$:0.72 $F_1$:0.72 |
| User4-1 | A:0.79 $F_0$:0.77 $F_1$:0.81 | A:0.76 $F_0$:0.76 $F_1$:0.75 | A:0.97 $F_0$:0.97 $F_1$:0.98 | A:0.95 $F_0$:0.96 $F_1$:0.91 | A:0.86 $F_0$:0.83 $F_1$:0.88 | A:0.91 $F_0$:0.89 $F_1$:0.92 |  | **A:0.62** $F_0$**:0.60** $F_1$**:0.63** | A:0.93 $F_0$:0.91 $F_1$:0.94 | A:0.91 $F_0$:0.90 $F_1$:0.93 |
| User5-1 | A:0.82 $F_0$:0.85 $F_1$:0.79 | A:0.86 $F_0$:0.87 $F_1$:0.83 | A:0.91 $F_0$:0.91 $F_1$:0.90 | A:0.92 $F_0$:0.92 $F_1$:0.91 | A:0.70 $F_0$:0.73 $F_1$:0.65 | ***A:0.63*** ***$F_0$:0.66*** ***$F_1$:0.59*** | A:0.88 $F_0$:0.90 $F_1$:0.84 | A:0.92 $F_0$:0.93 $F_1$:0.90 |  | **A:0.65** $F_0$**:0.67** $F_1$**:0.62** |

Sessions simulating users typing on different keyboards were all recognized as different with A > 0.9, including those from the same agent.

In Figure 3, 4 and 5 we have a detailed summary of a few analysis including a graphical representation of p values, the relative importance of each feature and a comparison of the distribution for both sessions under analysis for the most important feature.

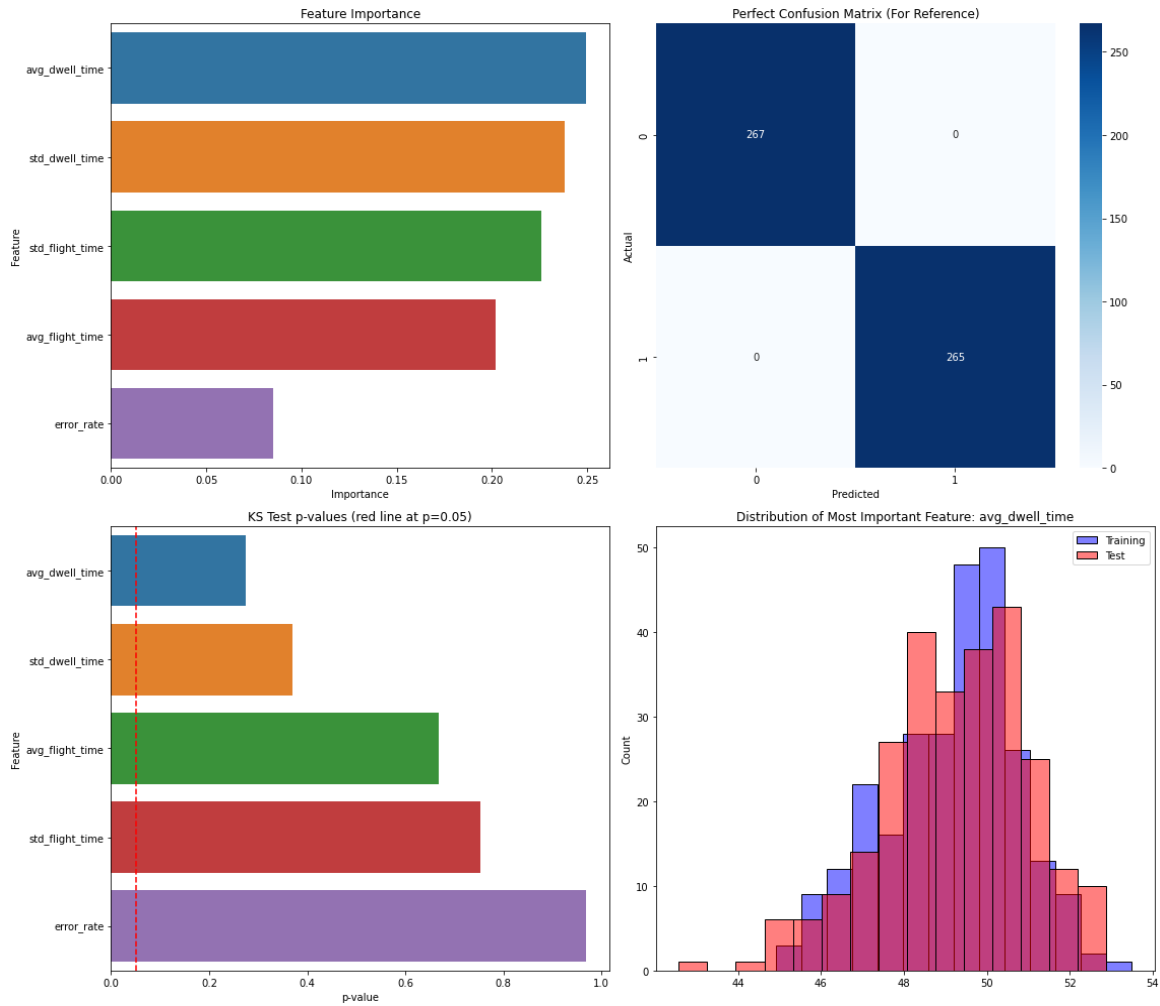

**Figure 3.** Analysis for User 1-1 vs User 1-2 sessions on a laptop keyboard, which are correctly recognized by the Random Forest analysis as belonging to the same user. Average Dwell time was identified as the most important feature.

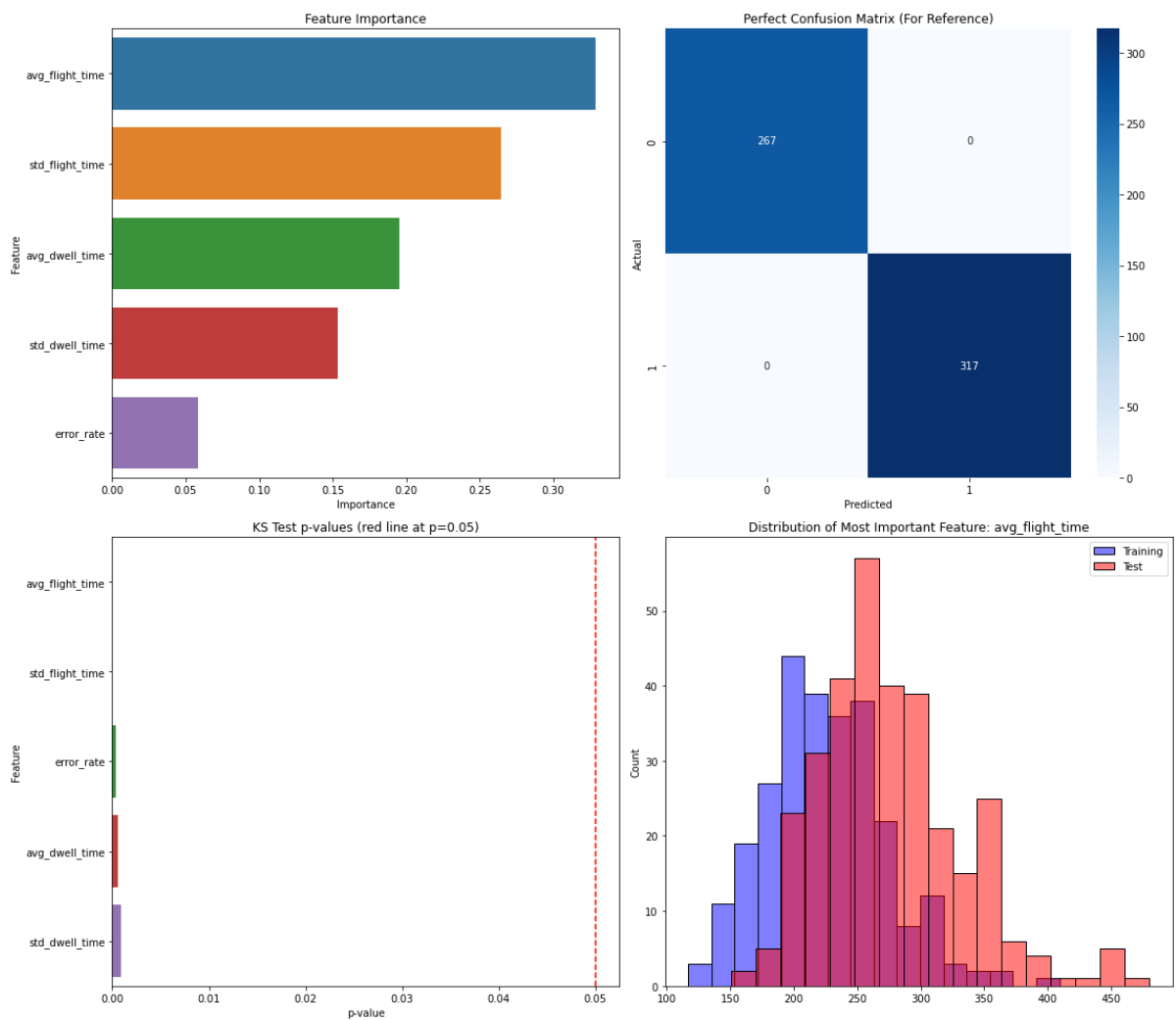

**Figure 4.** Analysis for User 1-1 vs User 3-2 sessions on a laptop keyboard, which are correctly recognized by the Random Forest analysis as belonging to the different user. Average Flight time was identified as the most important feature.

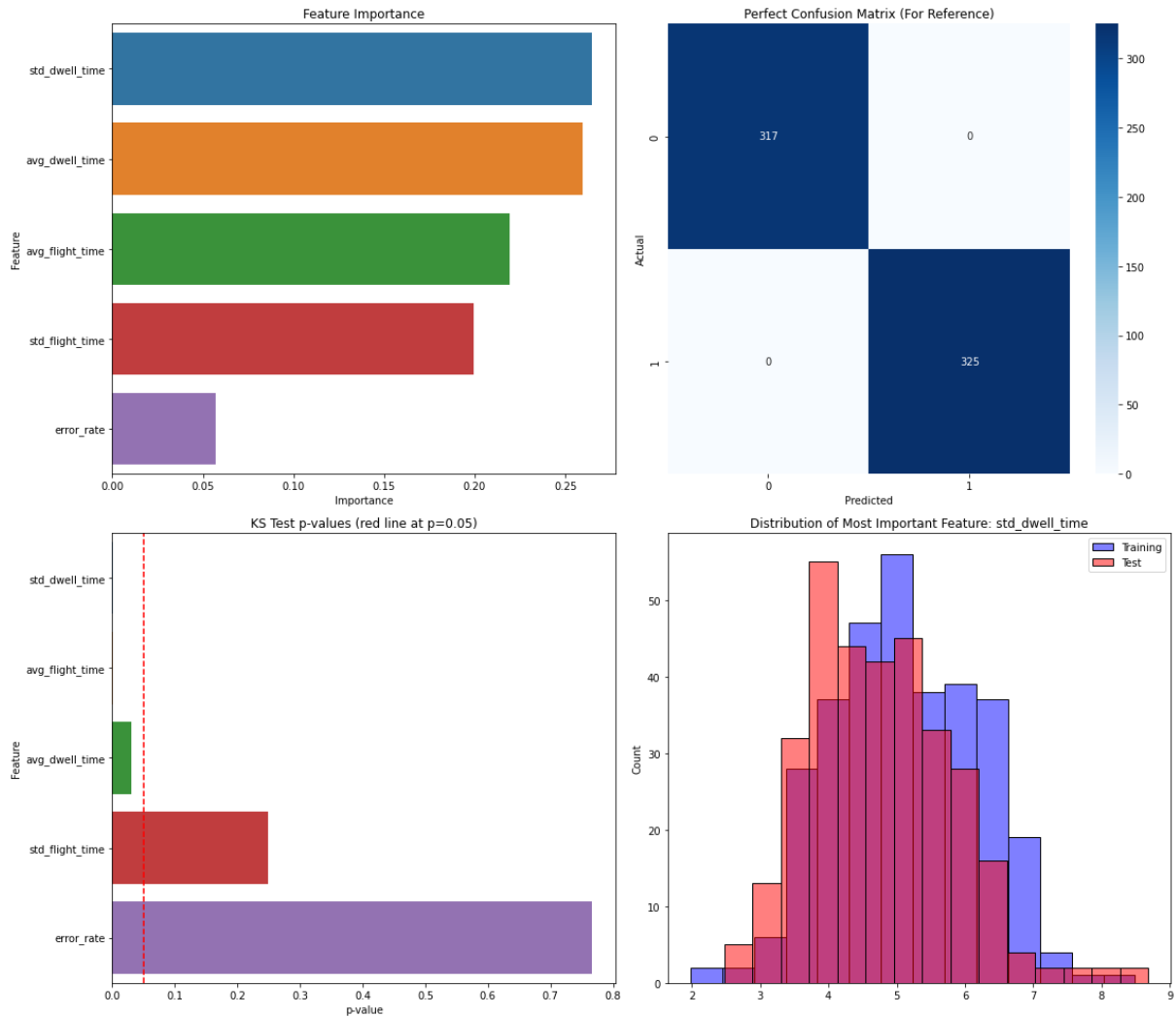

**Figure 5.** Analysis for User 3-1 vs User 5-2 sessions on a laptop keyboard, which are wrongly identified by the Random Forest as belonging to the same user. Standard deviation of the dwell time was identified as the most important feature and, despite three features still scoring low p values, the overlapping of error rate and standard deviation of the flight time are enough to confuse the recognition system.

## 6. Discussion

The OC-SVM model was able to show some promising results with some same-user cases (User 4, User 5, User 2), same session scenarios, but its performance across different users remained inconsistent. In fact, the OC-SVM did not consistently distinguish between different users and, even though the KS tests indicated statistical differences across the user typing behaviors, the OC-SVM's performance in cross-user data samples indicated limited differences and separability. In cross-user evaluations, the model yielded relatively high inlier detection rates (often exceeding 60%). Distributions like those plotted in Figures 1 and 2 were common across the simulations and indicated that test samples from other users frequently fell within the learned decision boundary, hence misclassifying a large portion of the test data as inliers and producing false negatives. These misclassifications highlight the spread of user distributions across the PCA space, suggesting that the first two components may not retain sufficient discriminatory information between users. These results highlight a limitation of the OC-SVM approach in its current configuration and suggest the need for either more informative features, additional PCA components, or stricter model hyperparameters (e.g., increased) to improve user separability.

On the other hand, the RF model demonstrated strong performance in differentiating between distinct simulated users, with accuracy scores (A) predominantly exceeding 0.7 for comparisons involving different users. For instance, comparisons such as U1-1 vs. U2-1 (A=0.88 for laptop, A=0.91 for mechanical) and U3-1 vs. U4-1 (A=0.93 for laptop, A=0.87 for mechanical) consistently achieved high accuracy and F1 scores. This suggests that the model reliably identifies typing sessions from different agents, regardless of the keyboard type. The model also successfully recognized typing sessions from the same simulated user, as indicated by accuracy scores below the 0.7 threshold (e.g., U1-1 vs. U1-2: A=0.66 for laptop, A=0.68 for mechanical). These results align with the expectation that sessions from the same user would exhibit similar typing dynamics, making them harder to distinguish.

Some cases were more challenging, though, and there were a few instances of misclassification where sessions from different users were incorrectly identified as coming from the same user instead. "User 3", in particular, was the most difficult profile to analyze. Nonetheless, it is interesting to point out how, on the laptop keyboard simulation, all four cross tests between User 3 and User 5 yielded misclassification results while, on the mechanical keyboard-based simulation, only one was misclassified (U5-1 vs U3-2) while the others still managed to pass the Accuracy threshold. These cases suggest that it can indeed be possible for certain users to exhibit overlapping typing patterns, potentially due to similar typing styles or other confounding factors underscoring the need for further refinement of the model or additional features to improve discrimination in such scenarios and avoid resulting false negatives.

To assess the model overall performance, it is also important to evaluate the F1 scores for both Class 0 and Class 1. These were generally balanced, indicating that the model does not exhibit significant bias toward either class. This balance is crucial for ensuring reliable performance in real-world applications, such as user identification, where both false positives and false negatives carry important consequences.

While results were consistent across keyboards, with the model showing no significant degradation in accuracy or F1 scores between the two types, the model was unable to recognize the same simulated user across different keyboards, showing the ABM model was not able to generalize a unique typing style across different conditions. Interestingly, this is consistent with the findings reported in [12] affirming that only skilled users are consistent across different keyboard types, and only one training profile may be needed for them. On the contrary, average or not-so-skilled users may instead showcase significant differences in their typing behaviors and individual profiles may be needed depending on the specific equipment used.

## 7. Conclusion

The ability to conduct ABM simulations allowed us to test different models quickly and effectively, producing results that, while mixed, were very interesting for gaining relevant insights into the design and implementation of systems for continuous user authentication. For example, the Random Forest (RF) analysis of simulated sessions on laptop and mechanical keyboards confirmed the viability of free-text keyboard dynamics as a biometric identifier. The findings highlight the algorithm's effectiveness in distinguishing between different users while also revealing some challenges in certain scenarios where the misclassification cases highlighted the need for further research, possibly incorporating additional features, such as specific error patterns, to enhance the model's discriminatory power. However, the inability to recognize users across devices underscores a critical limitation for real-world deployment, where users often switch keyboards as they may work across offices and setups, meaning that specific user's profiles for each setup may be needed to possibly reduce the number of false positives.

# References

1. Vagal V, Dillon R. Reducing cyber risk in remote working. In: Kuah A, Dillon R, editors. Digital transformation in a post-COVID world. Boca Raton: CRC Press; 2021. p. 155–170. https://doi.org/10.1201/9781003148715-8.
2. Gunetti D, Picardi C. Keystroke analysis of free text. *ACM Trans Inform Syst Secur*. 2005;8(3):312–47. https://doi.org/10.1145/1085126.1085129.
3. Bryan WL, Harter N. Studies on the telegraphic language. The acquisition of a hierarchy of habits. *Psychol Rev*. 1899;6(4):345–75. https://doi.org/10.1037/h0073117.
4. Obaidat MS, Sadoun B. Verification of computer users using keystroke dynamics. *IEEE Trans Syst Man Cybern B Cybern*. 1997;27(2):261–9. https://doi.org/10.1109/3477.558812.
5. Dowland P, Furnell S, Papadaki M. Keystroke analysis as a method of advanced user authentication and response. In: IFIP TC11 17th International Conference on Information Security: Visions and Perspectives; 2002. p. 215–26.
6. Ahmed AA, Traore I. Biometric recognition based on free-text keystroke dynamics. *IEEE Trans Cybern*. 2014;44(4):458–72. https://doi.org/10.1109/tcyb.2013.2257745
7. Shadman R, Wahab AA, Manno M, Lukaszewski MS, Hou D, Hussain F. Keystroke dynamics: concepts, techniques, and applications. *arXiv*. 2023. arXiv:2303.04605
8. Dillon R. Who's typing? An experiment on keyboard dynamics for BEC detection. In: *Proc 15th Int Workshop Appl Modelling Simulation*; 2024. p. 29–34. *https://www.liophant.org/conferences/2024/wams/papers/WAMS_221.pdf*
9. English letter frequencies, Corpus of Contemporary American English. https://www.wordfrequency.info/coca.asp
10. Feit AM, Weir D, Oulasvirta A. How we type: movement strategies and performance in everyday typing. In: Proc ACM Conf Human Factors Comput Syst (CHI); 2016.
11. Zeid S, ElKamar RA, Hassan IS. Fixed-text vs. free-text keystroke dynamics for user authentication. Eng Res J (Shoubra). 2022;51(1):95–104. https://doi.org/10.21608/erjsh.2022.224312
12. Matsubara Y, Samura T, Nishimura H. Keyboard dependency of personal identification performance by keystroke dynamics in free text typing. J Inf Secur. 2015; 6:229–40. https://doi.org/10.4236/jis.2015.63023.
16